# A Node-collaboration-informed Graph Convolutional Network for Precise Representation to Undirected Weighted Graphs

Ying Wang, Ye Yuan, Xin Luo

*Abstract*—An undirected weighted graph (UWG) is frequently adopted to describe the interactions among a solo set of nodes from real applications, such as the user contact frequency from a social network services system. A graph convolutional network (GCN) is widely adopted to perform representation learning to a UWG for subsequent pattern analysis tasks such as clustering or missing data estimation. However, existing GCNs mostly neglects the latent collaborative information hidden in its connected node pairs. To address this issue, this study proposes to model the node collaborations via a symmetric latent factor analysis model, and then regards it as a node-collaboration module for supplementing the collaboration loss in a GCN. Based on this idea, a Node-collaboration-informed Graph Convolutional Network (NGCN) is proposed with three-fold ideas: a) Learning latent collaborative information from the interaction of node pairs via a node-collaboration module; b) Building the residual connection and weighted representation propagation to obtain high representation capacity; and c) Implementing the model optimization in an end-to-end fashion to achieve precise representation to the target UWG. Empirical studies on UWGs emerging from real applications demonstrate that owing to its efficient incorporation of node-collaborations, the proposed NGCN significantly outperforms state-of-the-art GCNs in addressing the task of missing weight estimation. Meanwhile, its good scalability ensures its compatibility with more advanced GCN extensions, which will be further investigated in our future studies.

## I. INTRODUCTION

RECENT years have witnessed the great success of graph convolutional network (GCN) [1] in representation learning for graph-structure data. Owing to its effectiveness in graph representation learning, a GCN has been applied to solve a plethora of real-world problems competently. For instance, James [8] proposes a novel spatial-temporal GCN, which adopts a spatial-temporal attention mechanism for traffic flow forecasting effectively. Yu *et al*. [9] propose an adversarial GCN, which adopts adversarial training for precise recommendation. Azadifar *et al*. [10] utilize a deep GCN for gene essentiality prediction.

An undirected weighted graph (UWG) is frequently encountered graph-structure data in real applications [6,11-13,15]. In particular, a UWG assigns a quantifiable index to the interaction between two nodes as its weight, which represents the correlation or strength of the interaction between a pair of nodes. For instance, the weight can describe the citation counts between authors [17] in citation network, the confidence of interactome between proteins [18] in a protein network, and the contact rate between users [19] in a social network. From this point of view, we clearly see that node-node interactions in a UWG contain massive useful knowledge to help us better understand inter-relationship among nodes. However, when addressing a UWG, a GCN heavily relies on node features without considering the latent collaborative information hidden in its connected node pairs, which is crucial for missing weight estimation [20]. Therefore, present empirical GCN show poor performance in missing weight estimation of a UWG, and is even outperformed by a linear representation learning model [21]. In line with the aforementioned discoveries, we have encountered the research question:

*RQ.* Is it possible to incorporate collaborative information learned from the interaction of node pairs, thereby improving the GCN's performance on missing weight estimation?

To answer this question, this study proposes a Collaborative-informed Graph Convolutional Network (NGCN), which can exploit the latent collaborative information from the interactions of node pairs effectively. Specifically, it consists of the following three-fold ideas: a) Learning latent collaborative information from the interaction of node pairs via a node-collaboration module, b) Building the residual connection and weighted representation propagation to obtain high representation capacity, and c) Implementing the model optimization in an end-to-end fashion to achieve precise representation to the target UWG.

✧ Y. Wang is with the School of Computer Science and Technology, Chongqing University of Posts and Telecommunications, Chongqing 400065, China, and also with the Chongqing Key Laboratory of Big Data and Intelligent Computing, Chongqing Engineering Research Center of Big Data Application for Smart Cities, and Chongqing Institute of Green and Intelligent Technology, Chinese Academy of Sciences, Chongqing 400714, China (e-mail: 15032382660@163.com).
✧ Y. Yuan and X. Luo are with the College of Computer and Information Science, Southwest University, Chongqing 400715, China (e-mail: yuanyekl@gmailswu.edu.cn.com, luoxin@swu.edu.cn).

This paper mainly contributes in the following perspectives: a) We propose an NGCN model. Different from popular GCNs that rely solely on smoothed node features, it is able to incorporate learnable collaborative information hidden in the interactions of node pairs to enhance the representation ability on a UWG, and b) The proposed NGCN are comprehensively investigated on extensive benchmark datasets demonstrate it outperforms the state-of-the-art baselines in missing weight estimation.

Section 2 states the preliminaries. Section 3 presents the proposed NGCN. Section 4 conducts the empirical studies. Section 5 is the related work. Finally, Section 6 concludes this paper.

## II. PRELIMINARIES

### A. Problem Formulation

Note that a UWG is defined as [25, 26]:

***Definition* 1:** Let $V=\{v_i \mid i=1, 2, \ldots, N\}$ represent a set of $N$ nodes, $E=\{e_{ij} \mid v_i, v_j \in V\}$ represent a set of edges where the nodes $v_i, v_j \in V$ are connected. In particular, for a UWG $G=(V, E)$, assigns a quantifiable index to the edge between two nodes as its weight. Hence, it is denoted the adjacency matrix $A=[a_{ij}] \in \mathbf{R}^{N \times N}$, where $a_{ij}$ is the non-zero weight if $e_{ij} \in E$ and zero otherwise. Each node of $G$ is described by a feature vector $x_i \in \mathbf{R}^f$, where $X \in \mathbf{R}^{N \times f}$ is the collection of such vectors in a matrix form.

The missing weight estimation for a UWG is defined as [21]:

***Definition 2:*** Given a UWG $G=(V, E)$, a representation learning model learns meaningful representation $h_i \in \mathbf{R}^f$ for each node $i$ by an iterative learning process, where $H \in \mathbf{R}^{N \times f}$ is the collection of such vectors in a matrix form. Subsequently, $H$ is used for predicting the link weight of each link $(i, j)$ as follows:

$$\hat{a}_{ij} = \langle h_i, h_j \rangle, \tag{1}$$

where $<\cdot,\cdot>$ denotes the inner product of two vectors, $\hat{a}_{ij}$ denotes the predicted weight. Note that we estimate the missing weight rather than the probability of link between two nodes. Thus, the non-liner activation function is omitted here [28, 29].

### B. Graph Convolutional Network

As described in prior studies [30,31], a GCN learns low dimensional representations of nodes via aggregation of features from neighbors using non-linear transformations. Its forward propagation for node representation is defined as:

$$H^{(l+1)} = \sigma\left(D^{-0.5}(A+I)D^{-0.5}H^{(l)}W^{(l)}\right) \tag{2}$$

where $H^{(0)}$ is the node feature matrix $X$, $H^{(l)}$ denotes the representation of nodes at $l$-th layer, $H^{(l+1)}$ denotes the output representation after stacking $l$ GCN layers, $W^{(l)} \in \mathbf{R}^{d(l) \times d(l+1)}$ is a trainable weight matrix, $I$ denotes the identity matrix, $D$ denotes the degree matrix of $(A+I)$, $\sigma(\cdot)$ denotes the activation function.

## III. METHODS

We now introduce the proposed NGCN model. It consists of the following three main components: a) The residual and weighted representation propagation module employs the residual connection and weighted representation propagation into GCN to offer better node representation capacity, b) The collaborative information learning module extracts the useful collaborative information from the interactions of node pairs via a symmetric latent factor analysis model, and c) The estimation module combines collaborative information and node representation to estimate the missing weight.

### A. Weight Representation Propagation Module

Given a UWG in ***Definition 1***, this module takes the node feature matrix $X$ and weighted adjacency matrix $A$ as the input. Following with the forward propagation process in (2), we have that:

$$H^{(0)} = X, \tag{3a}$$

$$\tilde{H}^{(l+1)} = \text{ReLU}\left(D^{-0.5}AD^{-0.5}H^{(l)}W^{(l)}\right), \tag{3b}$$

$$H^{(l+1)} = \tilde{H}^{(l+1)} + H^{(l)}, \tag{3c}$$

where we adopt ReLU as the activation function in this module, $\tilde{H}^{(l+1)}$ denotes the output of l-th layer, $H^{(l+1)}$ and $H^{(l)}$ denotes the input of $(l+1)$-th and $l$-th layer, respectively.

Note that in (3c), the residual connection is adopted by combining the input and output of the $l$-th layer to obtain the input representation of $(l+1)$-th layer. It can facilitate shallow layer's feature reuse and gradient back-propagation, thereby offering better representation capacity. Moreover, this operation considers the effect of self-connection. Hence, the self-connection in (3b) is neglected to avoid the numerical instabilities of representation scale.

In order to better understanding (3a)-(3c), we reformulated them into a vector form. Thus, for node $i$, we have that:

$$h_i^{(0)} = x_i, \tag{4a}$$

$$\tilde{h}_i^{(l+1)} = \text{ReLU}\left(\left(\sum_{j \in N(i)} a_{ij} \frac{1}{\sqrt{d_{ii} d_{jj}}} h_j^{(l)}\right) W^{(l)}\right), \tag{4b}$$

$$h_i^{(l+1)} = \tilde{h}_i^{(l+1)} + h_i^{(l)}, \tag{4c}$$

where $N(i)$ denotes the directly-connected neighbor set of node $i$, $d_{ii}$ is the weighted diagonal degree of node $i$, $\tilde{h}_i^{(l+1)}$, $h_i^{(l+1)}$ and $h_i^{(l)}$ denote the $i$-th row vector corresponding to $\tilde{H}^{(l+1)}$, $H^{(l+1)}$ and $H^{(l)}$, respectively.

Note that the task of this study is missing weight estimation. Hence, (4a)-(4c) denotes a weighted representation propagation process, which is different from that of traditional GCN designed for node classification and link estimation in the following aspects:

a) For (4a), the node feature $x_i$ is randomly initialized vector and needs to be trained; and
b) For (4b), $d_{ii}$ denotes sum of weights between node $i$ and its directly-connected neighbors instead of computing its number of neighbor $|N(i)|$.

### B. Collaborative Information Learning Module

As aforementioned, the proposed NGCN is able to make use of the latent collaborative information from the interactions of node pair, which can be learned outside of GCN. Next, we propose a simple but effective way for NGCN to capture latent collaborative information.

Let $S=[s_{ij}] \in \mathbf{R}^{N \times N}$ denote the collaborative information matrix, where its element $s_{ij}$ is the collaborative information between nodes $i$ and $j$ in a UWG. It can be achieved with the following generalized generating function:

$$s_{ij} = \arg\min_{\Psi(S)_{ij}} \Phi\left(a_{ij}, \Psi(S)_{ij}\right), \tag{5}$$

where $\Phi(\cdot)$ denotes a distance metric, $\psi(\cdot)$ denotes an operator to construct the approximation of $A=[a_{ij}]$.

In this study, we adopt symmetric latent factor analysis (SLFA) [21-24] to achieve this purpose. The main reason is that SLFA can extract latent collaborative information from the interactions of node pair accurately since it considers the intrinsic symmetry property of the target UWG. Following the principle of SLFA, let $\Phi(\cdot)$ be the Euclidean distance-based metric and $\psi(S)_{ij} = (YY^T)_{ij}$, we have that:

$$s_{ij} = \arg\min_{(YY^T)_{ij}} \left(a_{ij} - (YY^T)_{ij}\right)^2, \tag{6}$$

where we use a latent matrix $Y \in \mathbf{R}^{N \times d}$ to approximate $S$ to reduce the computational burden, $d$ denotes the latent dimension of $Y$. Note that $s_{ij}$ is encoded via $(YY^T)_{ij}$ and it indicates that a higher collaboration level $s_{ij}$ learned by (6) means that nodes $i$ and $j$ are more similar. It could lead to better node representation in GCN.

### C. Estimation Module

Based on the inferences given in Sections 3.1 and 3.2, NGCN combines collaborative information and node representation to estimate the missing weight. NGCN computes the link weight $\hat{a}_{ij}$ as follows:

$$\hat{a}_{ij} = \omega \cdot \langle h_i, h_j \rangle + (1-\omega) \cdot \langle y_i, y_j \rangle, \tag{7}$$

where $y_i$ and $y_j$ denote the $i$-th and $j$-th row vector of $Y$, $\omega$ is trainable parameter to balance the importance of collaborative information and node representation.

In order to inject the collaborative information in an end-to-end fashion, we jointly train NGCN by considering the loss of each module. In this study, we utilize the commonly-adopted Euclidean distance to design the objective function:

$$L = \sum_{a_{ij} \in \Lambda} \left(a_{ij} - \hat{a}_{ij}\right)^2 + \sum_{a_{ij} \in \Lambda} \left(a_{ij} - \langle h_i, h_j \rangle\right)^2 + \sum_{a_{ij} \in \Lambda} \left(a_{ij} - \langle y_i, y_j \rangle\right)^2 + \lambda \|\Theta\|^2, \tag{8}$$

where $\Lambda$ denotes the training dataset, $\lambda$ denotes the regularization coefficient to control the $L_2$ regularization strength to prevent overfitting, $\Theta=\{\omega, X, Y, \{W\}_{l=1}^{L}\}$ denotes all trainable model parameters. We adopt mini-batch Adam [33] to optimize the estimation model and update the model parameters

## IV. Empirical Studies

### A. General Settings

**Evaluation Protocol**. This paper concerns the missing weight estimation for a UWG. Hence, we adopt the estimation accuracy as the evaluation protocol [4,34-36,38,39]. Commonly, root mean squared error (RMSE) and mean absolute error (MAE) [5,7,32,37,40,41] are commonly adopted to measure a model's estimation accuracy:

$$RMSE = \sqrt{\left(\sum_{a_{ij} \in \Gamma} a_{ij} - \hat{a}_{ij}\right)^2 / |\Gamma|}, MAE = \left(\sum_{a_{ij} \in \Gamma} |a_{ij} - \hat{a}_{ij}|_{abs}\right) / |\Gamma|$$

where |·| calculates the cardinality of an enclosed set, $|\cdot|_{abs}$ denotes the absolute value of an enclosed number, and $\Gamma$ denotes testing dataset.

**Datasets.** Four UWG datasets [42] are adopted in our experiments, whose details are summarized in Table 1. Note that we randomly split the known edges set of each UWG dataset into ten disjoint and equally-sized subsets, where seven subsets are chosen as the training set, one as the validation set, and the remaining two as the testing set for 70%-10%-20% train-validation-test setting. The above process is sequentially repeated five times for five-fold cross-validation.

**Baselines**. To demonstrate the effectiveness, we compare our proposed NGCN with the state-of-the-art models for missing weight estimation. Table 2 records all the compared models.

Table 1: Experimental dataset details.

| No. | Name | Edges | Nodes | Density |
|---|---|---|---|---|
| D1 | plantsmargin_12NN | 25482 | 1600 | 1.00% |
| D2 | yaleB_10NN | 34808 | 2414 | 0.60% |
| D3 | MISKnowledgeMap | 57022 | 2427 | 0.96% |
| D4 | yeast_30NN | 63250 | 1484 | 2.87% |

Table 2: Details of compared models.

| No. | Model | Description |
|---|---|---|
| M1 | MF | A matrix factorization-based model [43]. |
| M2 | NeuMF | A deep neural network-based MF model [44]. |
| M3 | GCN | A standard graph convolutional network [1]. |
| M4 | LR-GCCF | A linear residual GCN [45]. |
| M5 | DGCN_HN | A deep GCN with hybrid normalization [46]. |
| M6 | GCMC | A graph auto-encoder network [47]. |
| M7 | LightGCN | A Simplifying GCN [48]. |
| M8 | NGCN | The proposed NGCN of this study. |

**Training Settings**. For achieving the objective results, following general settings are applied to all involved models:
a) All the compared models are deployed on a GPU platform with two NVIDIA GeForce RTX 3090 GPU cards;
b) We adopt an Adam optimizer and the batch-size is set as 2048;
c) The termination conditions are consistent for all the compared models, i.e., the iteration threshold is 1000, or the training will be terminated after the minimum error of 30 iterations;
d) The node feature matrix of each model are initialized with the same randomly generated arrays, and the dimension of node feature $f$=128;
e) For the proposed NGCN, the latent matrix's latent dimension $d$=128, the number of propagation layers $L$=2;
f) For all the compared models, we apply a grid search for the learning rate $\eta$={0.00005, 0.0001, 0.0005, 0.001, 0.005, 0.01, 0.05} and $L_2$ regularization coefficient $\lambda$={0.00001, 0.00005, 0.0001, 0.0005, 0.001, 0.005, 0.01, 0.05, 0.1, 0.5, 1} to achieve the optimal results.

### B. Comparison Performance

We start by comparing the performance of all the methods. Table 3 show the comparison results on RMSE/MAE. From them, we achieve some important verdicts:

a) **NGCN possesses remarkable advantage in missing weight estimation of a UWG**. As depicted in Table 3, M8, i.e., the proposed NGCN, achieves the lowest estimation errors on seven testing cases out of eight in total. For instance, on D1, M8 achieves the optimal RMSE at 0.03520, which is 14.63% lower than 0.04123 achieved by M1, 8.19% lower than 0.03834 achieved by M2, 8.52% lower than 0.03848 achieved by M3, 12.57% lower than 0.04026 achieved by M4, 10.64% lower than 0.03939 achieved by M5, 7.81% lower than 0.03818 achieved by M6, and 8.74% lower than 0.03857 achieved by M7. Similar outcomes are also found on MAE. The situation is a little different on RMSE of D2. For this case, M8 is only slightly outperformed by M7. One possible interpretation is that D2 contains valuable linear characteristics. In general, M8's representation ability to the target UWG is impressive.

Table 3: The comparison results on RMSE/MAE, where ✪ indicates that NGCN is outperformed by the compared models.

| Case | | M1 | M2 | M3 | M4 | M5 | M6 | M7 | M8 |
|---|---|---|---|---|---|---|---|---|---|
| D1 | RMSE | $0.04123_{\pm 5.E-5}$ | $0.03834_{\pm 2.E-4}$ | $0.03848_{\pm 2.E-4}$ | $0.04026_{\pm 2.E-3}$ | $0.03939_{\pm 3.E-4}$ | $0.03818_{\pm 1.E-4}$ | $0.03857_{\pm 5E-5}$ | $\mathbf{0.03520_{\pm 8.E-5}}$ |
| | MAE | $0.02312_{\pm 1.E-4}$ | $0.02317_{\pm 5.E-4}$ | $0.02320_{\pm 8.E-5}$ | $0.02298_{\pm 9.E-5}$ | $0.02171_{\pm 3.E-5}$ | $0.02280_{\pm 4.E-4}$ | $0.02168_{\pm 1.E-5}$ | $\mathbf{0.02068_{\pm 2.E-4}}$ |
| D2 | RMSE | $0.07388_{\pm 1.E-4}$ | $0.07029_{\pm 5.E-4}$ | $0.06921_{\pm 8.E-4}$ | $0.07205_{\pm 8.E-4}$ | $0.07046_{\pm 1.E-5}$ | ✪$0.06796_{\pm 4.E-4}$ | $0.06912_{\pm 8.E-5}$ | $0.06873_{\pm 2.E-4}$ |
| | MAE | $0.03445_{\pm 4.E-5}$ | $0.03508_{\pm 6.E-4}$ | $0.03504_{\pm 9.E-4}$ | $0.03774_{\pm 6.E-4}$ | $0.03336_{\pm 3.E-5}$ | $0.03595_{\pm 5.E-4}$ | $0.03341_{\pm 5.E-5}$ | $\mathbf{0.03191_{\pm 8.E-5}}$ |
| D3 | RMSE | $0.07355_{\pm 7.E-5}$ | $0.06121_{\pm 2.E-4}$ | $0.07339_{\pm 6.E-4}$ | $0.09000_{\pm 2.E-3}$ | $0.07453_{\pm 2.E-4}$ | $0.06238_{\pm 2.E-4}$ | $0.07614_{\pm 3.E-4}$ | $\mathbf{0.06050_{\pm 4.E-4}}$ |
| | MAE | $0.05030_{\pm 1.E-4}$ | $0.04527_{\pm 5.E-4}$ | $0.05564_{\pm 6.E-4}$ | $0.06150_{\pm 1.E-3}$ | $0.05062_{\pm 2.E-5}$ | $0.04651_{\pm 3.E-4}$ | $0.05030_{\pm 4.E-5}$ | $\mathbf{0.04368_{\pm 5.E-4}}$ |
| D4 | RMSE | $0.08809_{\pm 1.E-4}$ | $0.08619_{\pm 2.E-4}$ | $0.08576_{\pm 1.E-3}$ | $0.08716_{\pm 3.E-4}$ | $0.08203_{\pm 4.E-5}$ | $0.08231_{\pm 7.E-4}$ | $0.08141_{\pm 5.E-5}$ | $\mathbf{0.07323_{\pm 2.E-4}}$ |
| | MAE | $0.04523_{\pm 4E-5}$ | $0.05366_{\pm 8.E-4}$ | $0.05181_{\pm 9.E-4}$ | $0.05046_{\pm 4.E-4}$ | $0.04227_{\pm 5.E-5}$ | $0.04796_{\pm 4.E-4}$ | $0.04224_{\pm 3.E-5}$ | $\mathbf{0.04028_{\pm 2.E-4}}$ |
| ✪Loss/Win | | 0/8 | 0/8 | 0/8 | 0/8 | 0/8 | 1/7 | 0/8 | — |

Compared with the MF-based models, i.e., M1 and M2, M8 show significant accuracy improvements. The main reason is that M8 incorporates the information of neighbors by the weighted representation forward propagation process, thereby improving the representation learning to the target UWG. Moreover, compared with the GCN-based models, i.e., M3-M7, M8 not only utilizes the information of neighbors, but also captures collaborative information from the interactions of node pairs, which is crucial for missing weight estimation. Hence, M8 still outperforms them when estimating the missing weight.

b) **NGCN's performance gain has statistical significance across our experiments**. To illustrate this point, the statistical analyses, i.e., the Friedman test and Wilcoxon signed-rank test, are conducted. Firstly, the Friedman test is commonly adopted to validate multiple models' performance on multiple datasets. The Friedman statistical results are recorded in Table 4, where it accepts the hypothesis that involved models have significant differences with a significance level of 0.05. From it, we clearly see that M8, i.e., the proposed NGCN, achieves the lowest Rank value, which means it outperforms other compared models in terms of estimation accuracy.

In addition, the Wilcoxon signed-ranks test is effective in checking whether NGCN has a significantly better performance than each compared model. Note that three indicators are adopted in the Wilcoxon signed-ranks test, where large $R+$ value denotes performance gain, large $R-$ denotes performance loss, and small p-value denotes high significance level. Table 5 records the corresponding results, which demonstrates that NGCN's estimation accuracy for missing weight of a UWG is significantly higher than that of its peers.

Table 4: Results of the Friedman test.

| Rank | M1 | M2 | M3 | M4 | M5 | M6 | M7 | M8 |
|---|---|---|---|---|---|---|---|---|
| Accuracy | 6.0 | 4.9 | 5.5 | 7.0 | 4.4 | 3.6 | 3.5 | **1.1** |

Table 5: Results of the Wilcoxon Signed-rank test.

| Comparison | Accuracy | | |
|---|---|---|---|
| | $R+$ | $R-$ | *p*-value* |
| M8 vs M1 | 36 | 0 | **0.003906** |
| M8 vs M2 | 36 | 0 | **0.003906** |
| M8 vs M3 | 36 | 0 | **0.003906** |
| M8 vs M4 | 36 | 0 | **0.003906** |
| M8 vs M5 | 36 | 0 | **0.003906** |
| M8 vs M6 | 35 | 1 | **0.007813** |
| M8 vs M7 | 36 | 0 | **0.003906** |

\* The accepted hypotheses with a significance level of 0.05 are highlighted.

## V. RELATED WORK

To date, diverse GCNs [30,31, 49] have been proposed. For instance, Abu-El-Haija et al. [30] propose a mixhop GCN, which repeatedly mixes feature representations of neighbors at various distances. Pei *et al*. [31] propose a geometric GCN, which maps a graph to a continuous latent space via node representations, and then use the geometric relationships defined in the latent space to build structural neighborhoods for aggregation. Bo *et al*. [49] propose a deep structural GCN, which combines multiple structures from low-order to high-order. Fatemi *et a*l. [16] propose a self-supervision GCN, which can provide more supervision for inferring a graph structure through self-supervision. Due to the iterative aggregation step, these noteworthy GCNs can achieve node representations for various downstream tasks. Obviously, estimating the missing weight of a UWG [2,20-23] is a typical downstream task, which is focused in this paper. However, when performing missing weight estimation, the existing GCNs neglect the latent collaborative information hidden in the interactions of node pairs, which is crucial for missing weight estimation.

According to prior researches [3,14,21-24,27], SLFA can learn the latent collaborative information from the concerned interaction of node pair, thereby implementing a precise representation of symmetry topology and numerical characteristics of a UWG. For instance, Li *et al*. [21] propose a second-order SLFA model, which designs an efficient second-order learning algorithm to achieve collaborative information with affordable computational burden. Luo *et al*. [22] propose an Alternating-Direction-Method of Multipliers (ADMM)-based SLFA model, which incorporates the ADMM principle into its learning scheme for fast and accurate extraction of collaborative information. Song *et al*. [23] propose an improved SLFA model, which adopts a triple factorization technique to achieve collaborative information with high precision.

## VI. CONCLUSIONS

In this paper, we propose a Collaborative-informed Graph Convolutional Network (NGCN) that learns latent collaborative information from the interactions of node pairs outside of the GNN. It offers flexible incorporation of both node features and latent collaborative information of node pairs, thereby achieving significant performance improvements on missing weight estimation of a UWG.

In future, we will further improve the effectiveness of NGCN by incorporating attention mechanism to measure the importance of latent collaborative information for different node-pairs. Moreover, we plan to theoretically analyze why latent collaborative information of node pairs can help GCN break the limitation of expressive power, whose upper-bounded is proven by the 1-Weisfeiler-Lehman (1-WL) graph isomorphism test.

## REFERENCES

[1] M. Welling and T. N. Kipf, "Semi-supervised classification with graph convolutional al networks," in *Proc. Int. Conf. on Learning Representations*, 2016.
[2] D. Wu and X. Luo, "Robust Latent Factor Analysis for Precise Representation of High-Dimensional and Sparse Data," *IEEE/CAA J. Autom. Sinica*, vol. 8, no. 4, pp. 796-805, 2021.
[3] X. Luo, J. Sun, Z. Wang, S. Li and M. Shang, "Symmetric and Nonnegative Latent Factor Models for Undirected, High-Dimensional, and Sparse Networks in Industrial Applications," in *IEEE Trans. on Industrial Informatics*, vol. 13, no. 6, pp. 3098-3107, 2017.
[4] X. Luo, Y. Yuan, S. Chen, N. Zeng and Z. Wang, "Position-Transitional Particle Swarm Optimization-Incorporated Latent Factor Analysis," in *IEEE Transactions on Knowledge and Data Engineering*, vol. 34, no. 8, pp. 3958-3970, 2022.
[5] X. Luo, W. Qin, A. Dong, K. Sedraoui and M. Zhou, "Efficient and High-quality Recommendations via Momentum-incorporated Parallel Stochastic Gradient Descent-Based Learning," in *IEEE/CAA Journal of Automatica Sinica*, vol. 8, no. 2, pp. 402-411, 2021.
[6] L. Hu, S. Yang, X. Luo, H. Yuan, K. Sedraoui and M. Zhou, "A Distributed Framework for Large-scale Protein-protein Interaction Data Analysis and Prediction Using MapReduce," in *IEEE/CAA Journal of Automatica Sinica*, vol. 9, no. 1, pp. 160-172, 2022.
[7] L. Xin, Y. Yuan, M. Zhou, Z. Liu and M. Shang, "Non-Negative Latent Factor Model Based on β-Divergence for Recommender Systems," in *IEEE Transactions on Systems, Man, and Cybernetics: Systems*, vol. 51, no. 8, pp. 4612-4623, 2021.
[8] S. Guo, Y. Lin, N. Feng, C. Song, and H. Y. Wan, "Attention based spatial-temporal graph convolutional al networks for traffic flow forecasting," in *Proc. of the AAAI Conf. on Artificial Intelligence*, pp. 922-929, 2019.
[9] J. L. Yu, H. Z. Yin, J. D. Li, M. Gao, Z. Huang, and L. Z. Cui, "Enhancing Social Recommendation With Adversarial Graph Convolutional al Networks," *IEEE Trans. on Knowledge and Data Engineering*, vol, 34, no. 8, pp. 3727-3739, 2022.
[10] S. Azadifar and A. Ahmadi, "A novel candidate disease gene prioritization method using deep graph convolutional al networks and semi-supervised learning," *BMC bioinformatics*, vol. 23, no. 1, pp. 1-25, 2022.
[11] L. Hu, J. Zhang, X. Pan, X. Luo and H. Yuan, "An Effective Link-Based Clustering Algorithm for Detecting Overlapping Protein Complexes in Protein-Protein Interaction Networks," in *IEEE Trans. on Network Science and Engineering*, vol. 8, no. 4, pp. 3275-3289, 2021.
[12] F. He, W. C. Lee, T. Y. Fu, and Z. Lei, "CINES: Explore Citation Network and Event Sequences for Citation Forecasting," in *Proc. of the Int. ACM Conf. on Research and Development in Information Retrieval*, pp. 798-807, 2021.
[13] L. Hu, X. Yuan, X. Liu, S. Xiong and X. Luo, "Efficiently Detecting Protein Complexes from Protein Interaction Networks via Alternating Direction Method of Multipliers," in *IEEE/ACM Trans.on Computational Biology and Bioinformatics*, vol. 16, no. 6, pp. 1922-1935, 2019.
[14] X. Luo, Z. Liu, L. Jin, Y. Zhou and M. Zhou, "Symmetric Nonnegative Matrix Factorization-Based Community Detection Models and Their Convergence Analysis," in *IEEE Trans. on Neural Networks and Learning Systems*, vol. 33, no. 3, pp. 1203-1215, 2022.
[15] Z. -H. You, M. Zhou, X. Luo and S. Li, "Highly Efficient Framework for Predicting Interactions Between Proteins," in *IEEE Trans. on Cybernetics*, vol. 47, no. 3, pp. 731-743, 2017.

[16] B. Fatemi, L. E. Asri, and S. M. Kazemi, "SLAPS: Self-supervision improves structure learning for graph neural networks," in *Proc. of Advances in Neural Information Processing Systems*, pp. 22667-22681, 2021.
[17] D. Yu and T. Pan, "Tracing knowledge diffusion of TOPSIS: A historical perspective from citation network," *Expert Systems with Applications*, vol. 168, no. 2, pp. 114238, 2020.
[18] M. Hofree, J. P. Shen, H. Carter, A. Gross, and T. Ideker, "Network-based stratification of tumor mutations," *Nature Methods*, vol. 10, no. 11, pp.1108-1115, 2013.
[19] J. Zhang, C. Wang, and J. Wang, "Who proposed the relationship?: recovering the hidden directions of undirected social networks," in *Proc. Of Int. World Wide Web Conference*, pp.807-818, 2014.
[20] X. Luo, H. Wu, Z. Wang, J. J. Wang, and D. Y. Meng, "A Novel Approach to Large-Scale Dynamically Weighted Directed Network Representation," *IEEE Trans. on Pattern Analysis and Machine Intelligence*, DOI: 10.1109/TPAMI.2021.3132503, 2021.
[21] W. L. Li, R. F. Wang, X. Luo, and M. C. Zhou, "A Second-order Symmetric Non-negative Latent Factor Model for Undirected Weighted Network Representation," *IEEE Trans. on Network Science and Engineering*, DOI: 10.1109/TNSE.2022.3206802, 2022.
[22] X. Luo, Y. R. Zhong, Z. D. Wang, and M. Z. Li, "An Alternating-direction-method of Multipliers-Incorporated Approach to Symmetric Non-negative Latent Factor Analysis," *IEEE Trans. on Neural Networks and Learning Systems*, DOI: 10.1109/TNNLS.2021.3125774, 2021.
[23] Y. Song, M. Li, X. Luo, G. S. Yang and C. J. Wang, "Improved Symmetric and Nonnegative Matrix Factorization Models for Undirected, Sparse and Large-Scaled Networks: A Triple Factorization-Based Approach," *IEEE Trans on Industrial Informatics*, vol. 16, no. 5, pp. 2020.
[24] Z. Liu, G. Yuan and X. Luo, "Symmetry and Nonnegativity-Constrained Matrix Factorization for Community Detection," in *IEEE/CAA Journal of Automatica Sinica*, vol. 9, no. 9, pp. 1691-1693, 2022.
[25] R. Manríquez, C. Guerrero-Nancuante, F. Martínez, and C. Taramasco, "A generalization of the importance of vertices for an undirected weighted graph," *Symmetry*, vol. 13, no. 5, pp. 902, 2021.
[26] S. Pettie, and V. Ramachandran, "A shortest path algorithm for real-weighted undirected graphs," *SIAM Journal on Computing*, vol. 34, no. 6, pp. 1398-1431, 2005.
[27] X. Luo, Z. Liu, M. Shang, J. Lou and M. Zhou, "Highly-Accurate Community Detection via Pointwise Mutual Information-Incorporated Symmetric Non-Negative Matrix Factorization," in *IEEE Trans. on Network Science and Engineering*, vol. 8, no. 1, pp. 463-476, 2021.
[28] S. Yun, S. Kim, J. Y. Lee, J. Kang, H. J. Kim, "Neo-GNNs: Neighborhood Overlap-aware Graph Neural Networks for Link Prediction," in *Proc. of Advances in Neural Information Processing Systems*, pp. 13683-13694, 2021.
[29] M. Zhang and Y. Chen, "Link prediction based on graph neural networks," in *Proc. of Advances in Neural Information Processing Systems*, pp. 5171-5181, 2018.
[30] S. Abu-El-Haija, B. Perozzi, A. Kapoor, N. Alipourfard, K. Lerman, H. Harutyunyan, G. V. Steeg, and A. Galstyan, "Mixhop: Higher-order graph convolutional architectures via sparsified neighborhood mixing," in *Proc. Int. Conf. on Machine Learning*, pp. 21-29, 2019.
[31] H. B. Pei, B. Z. Wei, K. C. Chang, Y. Lei, and B. Yang, "Geom-GCN: Geometric Graph Convolutional Networks," in *Proc. Int. Conf. on Learning Representations*, 2020.
[32] X. Luo, Z. Wang and M. Shang, "An Instance-Frequency-Weighted Regularization Scheme for Non-Negative Latent Factor Analysis on High-Dimensional and Sparse Data," in *IEEE Trans. on Systems, Man, and Cybernetics: Systems*, vol. 51, no. 6, pp. 3522-3532, 2021.
[33] D. Kingma and J. Ba, "Adam: A Method for Stochastic Optimization," *arXiv:1412.6980*, 2014.
[34] Y. Yuan, X. Luo, M. S. Shang, and Z. D. Wang, "A Kalman-Filter-Incorporated Latent Factor Analysis Model for Temporally Dynamic Sparse Data," *IEEE Trans. on Cybernetics*, DOI: 10.1109/TCYB.2022.3185117, 2022.
[35] M. Shang, Y. Yuan, X. Luo and M. Zhou, "An α–β-Divergence-Generalized Recommender for Highly Accurate Predictions of Missing User Preferences," in *IEEE Trans. on Cybernetics*, vol. 52, no. 8, pp. 8006-8018, 2022.
[36] Y. Yuan, Q. He, X. Luo, and M. S. Shang, "A Multilayered-and-Randomized Latent Factor Model for High-Dimensional and Sparse Matrices," *IEEE Trans. on Big Data*, vol. 8, no. 3, pp. 784-794, 2022.
[37] H. Wu, X. Luo and M. Zhou, "Advancing Non-Negative Latent Factorization of Tensors With Diversified Regularization Schemes," in *IEEE Trans. on Services Computing*, vol. 15, no. 3, pp. 1334-1344, 2022.
[38] Y. Yuan, M. S. Shang and X. Luo, "Temporal Web Service QoS Prediction via Kalman Filter-Incorporated Latent Factor Analysis," *ECAI 2020. IOS Press*, pp. 561-568, 2020.
[39] Y. Yuan, X. Luo, M. Shang, and D. Wu, "A generalized and fast-converging non-negative latent factor model for predicting user preferences in recommender systems," *In Proc. of The Web Conf. 2020*, pp. 498-507, 2020.
[40] Y. Yuan, X. Luo, and M. S. Shang, "Effects of preprocessing and training biases in latent factor models for recommender systems," *Neurocomputing*, vol. 275, pp. 2019-2030, 2018.
[41] W. Li, Q. He, X. Luo and Z. Wang, "Assimilating Second-Order Information for Building Non-Negative Latent Factor Analysis-Based Recommenders," in *IEEE Trans. on Systems, Man, and Cybernetics: Systems*, vol. 52, no. 1, pp. 485-497, 2022.
[42] A. D. Timothy and Y. H. Hu, "The University of Florida sparse matrix collection," *ACM Trans. on Mathematical Software*, vol. 38, no. 1, pp. 1-25, 2001.
[43] X. Luo, D. X. Wang, M. C. Zhou, and H. Y. Yuan, "Latent factor-based recommenders relying on extended stochastic gradient descent algorithms," *IEEE Trans. on Systems, Man, and Cybernetics: Systems*, vol. 51, no. 2, pp. 916-926, 2021.
[44] X. N. He, L. Liao, H. Zhang, et al., "Neural collaborative filtering," in *Proc. of the Int. Conf. on World Wide Web*, pp. 173-182, 2017.
[45] L. Chen, L. Wu, R. Hong, K. Zhang, and M. Wang, "Revisiting graph based collaborative filtering: A linear residual graph convolutional network approach," in *Proc. of the AAAI Conf. on Artificial Intelligence*, pp. 27-34, 2020.
[46] W. Guo, Y. Yang, Y. C. Hu, C. Y. Wang, H. F. Guo, Y. X. Zhang, R. M. Tang, W. N. Zhang, X. Q. He, "Deep graph convolutional networks with hybrid normalization for accurate and diverse recommendation," in *Proc. of Workshop on Deep Learning Practice for High-Dimensional Sparse Data with KDD*, 2021.
[47] R. V. Berg, T. N. Kipf, and M. Welling, "Graph convolutional matrix completion," *arXiv preprint arXiv:1706.02263*, 2017.
[48] X. N. He, K. Deng, X. Wang, Y. Li, Y. D. Zhang, and M. Wang, "LightGCN: Simplifying and powering graph convolution network for recommendation," In *Proc. of the Int. ACM SIGIR Conf. on research and development in Information Retrieval*, pp. 639-648, 2020.
[49] D. Bo, X. Wang, C. Shi, M. Zhu, E. Lu, and P. Cui, "Structural deep clustering network," in *Proc. of the Int. Conf. on World Wide Web*, pp. 1400-1410, 2020.